\documentclass{article}

\usepackage{microtype}
\usepackage{graphicx}
\usepackage{subfigure}
\usepackage{booktabs} 
\usepackage{amsmath}
\usepackage{amssymb}
\usepackage{booktabs}
\usepackage{adjustbox}
\usepackage{stfloats}
\usepackage{amsthm}

\usepackage{hyperref}

\newtheorem{theorem}{Definition}

\usepackage[font=small,skip=0pt]{caption}

\usepackage[accepted]{icml2021}

\icmltitlerunning{Differentially private training of neural networks with Langevin dynamics for calibrated predictive uncertainty}

\begin{document}

\twocolumn[
\icmltitle{Differentially private training of neural networks with Langevin dynamics for calibrated predictive uncertainty}




\begin{icmlauthorlist}
\icmlauthor{Moritz Knolle}{rad,aim}
\icmlauthor{Alexander Ziller}{rad,aim,om}
\icmlauthor{Dmitrii Usynin}{aim,icl,om}
\icmlauthor{Rickmer Braren}{rad}
\icmlauthor{Marcus R. Makowski}{rad}
\icmlauthor{Daniel Rueckert}{aim,icl}
\icmlauthor{Georgios Kaissis}{rad,aim,om,icl}
\end{icmlauthorlist}

\icmlaffiliation{rad}{Department of Diagnostic and Interventional Radiology, Klinikum rechts der Isar, School of Medicine,  Technical University of Munich, Munich, Germany}
\icmlaffiliation{aim}{Institute for Artificial Intelligence and Informatics in Medicine, Klinikum rechts der Isar, School of Medicine, Technical University of Munich, Munich, Germany}
\icmlaffiliation{om}{OpenMined}
\icmlaffiliation{icl}{Department of Computing, Imperial College London, London, United Kingdom}

\icmlcorrespondingauthor{Georgios Kaissis}{g.kaissis@tum.de}

\icmlkeywords{Machine Learning, ICML}

\vskip 0.3in
]



\printAffiliationsAndNotice{\icmlEqualContribution} 

\begin{abstract}
We show that differentially private stochastic gradient descent (DP-SGD) can yield poorly calibrated, overconfident deep learning models. This represents a serious issue for safety-critical applications, e.g. in medical diagnosis. We highlight and exploit parallels between stochastic gradient Langevin dynamics, a scalable Bayesian inference technique for training deep neural networks, and DP-SGD, in order to train differentially private, Bayesian neural networks with minor adjustments to the original (DP-SGD) algorithm. Our approach provides considerably more reliable uncertainty estimates than DP-SGD, as demonstrated empirically by a reduction in expected calibration error (MNIST $\sim{5}$-fold, Pediatric Pneumonia Dataset $\sim{2}$-fold).
\end{abstract}

\section{Introduction}
Safety-critical applications of machine learning require calibrated predictive uncertainty, that is, neither over- or under-confident uncertainty estimates. However, modern trends in deep neural network (DNNs) architecture design and training such as strong overparameterization,  normalization,  and  regularization have been shown to have negative effects on calibration, yielding overconfident models \cite{guo2017calibrationNNs}. A natural solution is the application of Bayesian inference to DNNs, as it provides a sound framework for making optimal predictions under uncertainty \cite{mackay1992practical, ovadia2019can, wilson2020case}.

The successful use of DNNs for a variety of real-world problems and tasks\cite{mckinney2020international, silver2016mastering, poplin2018prediction}, has shown that the application of deep learning techniques to humanity's most important problems is mainly held back by lack of usable data rather than by technological immaturity. This is particularly evident in the medical domain, where the application of machine learning has hitherto been limited by small, single-institutional and often non-representative datasets, due to the sensitive nature of medical data and strict regulation governing its use. As a result, even large published studies \cite{mckinney2020international} have utilized datasets that are an order of magnitude smaller than commonly used computer vision datasets such as ImageNet \cite{deng2009imagenet}. Overconfident models trained on small, non-representative datasets are especially undesirable in domains such as medicine, where, contrary to generic computer vision tasks in which the quality of predictions is typically assessed in aggregate over a test set, decisions have direct, individual consequences for affected patients.

Two \textendash mutually complementary \textendash solutions, can be employed to address the above-mentioned issue: (1) The utilisation of privacy-enhancing and federated learning techniques allows drawing conclusions from data to which direct access is not possible, increasing the effective dataset size and diversity. (2) The application of Bayesian inference to neural networks allows for accurate uncertainty quantification and optimal decision making under uncertainty, resulting in better calibrated models.

In this work, we exploit the similarities between stochastic gradient Langevin dynamics (SGLD) \cite{welling2011bayesian}, a highly scalable stochastic gradient Markov Chain Monte Carlo (SG-MCMC) method and differentially private stochastic gradient descent (DP-SGD) \cite{abadi2016deep}, to provide formal privacy guarantees, while enabling more reliable and accurate uncertainty estimation.

\subsection{Main contributions}
\begin{itemize}
    \item We show that DP-SGD has a negative impact on 
    model calibration compared to standard SGD, making its deployment in safety-critical applications such as automated medical diagnosis problematic 
    \item We highlight and exploit similarities between DP-SGD and SGLD, to reformulate DP-SGD as a temperature-scaled SGLD. Our novel reformulation  offers more flexibility for achieving optimal privacy-utility trade-offs, by removing the need for a direct, 1:1 coupling of learning rate and noise scaling parameter
    \item We provide empirical evidence that DP-SGLD provides substantially better calibrated predictive uncertainty than standard DP-SGD
\end{itemize}

\section{Related Work}
    Prior work has established the natural link between differentially private training of neural networks and stochastic gradient Langevin dynamics \cite{wang2015privacy4free, li2019connecting}. Both however, do not take into account the effects of gradient clipping with regards to gradient bias. Moreover, step size decay is neglected in \cite{li2019connecting} for training of neural networks, a formal convergence requirement of SGLD which is often disregarded in practice. \textit{Guo et al.} \cite{guo2017calibrationNNs} showed that modern, very deep neural networks often provide miscalibrated uncertainty estimates and propose expected calibration error (ECE) as an empirical, approximate measure of miscalibration. ECE uses the maximum probability of the softmax output as a measure of model confidence to approximate the calibration error, however more recent work \cite{nixon2019measuringCalibration} identifies issues with this approach and introduces static calibration error (SCE), which measures calibration of all classes and adaptive calibration error (ACE), which picks bins for the approximation adaptively to contain similar amounts of samples (ACE). \textit{Post-hoc} (re)-calibration methods \cite{platt1999probabilistic, guo2017calibrationNNs} have been shown to significantly improve calibration  without hindering the performance. They, however, rely on the assumption that the validation set used for recalibration is fully representative of the target distribution. This assumption may be problematic as some distributional shift between validation set and target distribution is to be expected, as supported by empirical evidence \cite{ovadia2019can}.\newline
    So far, to the best of our knowledge, no prior work has analysed the benefits of utilising Langevin dynamics for private learning over standard DP-SGD with respect to model calibration.

\section{Background}
    \subsection{Differential privacy and differentially private deep learning}
        Differential privacy \cite{dwork2006our, dwork2014algorithmic} (DP) is an information-theoretic privacy definition providing an upper bound for the information gain from observing the output of an algorithm applied to a dataset.
    
        \begin{theorem}
            For some randomised \textit{algorithm} \textit{$A\colon D\xrightarrow{}Y$}, all subsets of its image $Y$, sensitive dataset \textit{$d$} and its neighbouring dataset \textit{$d'$}, we say that $A$ is $(\epsilon, \delta)$-differentially private if, for a (typically small) constant $\epsilon$ and $0\leq\delta<1$: 
            \begin{equation}
                P[\mathit{A(d) = y}] \leq e ^ \epsilon P[\mathit{A(d^{\prime}) = y}] + \delta
            \end{equation}
            Here neighbouring datasets \textit{$D$} and \textit{$D'$} differ by at most one record and $y \in Y' \subseteq Y$.
        \end{theorem}
        \subsubsection{Differentially private deep learning}
            \begin{theorem}
            Differentially private stochastic gradient descent \cite{abadi2016deep} (DP-SGD) is generalization of differential privacy in the context of deep learning training. In DP-SGD, mini-batch gradients $\nabla_{b} \theta$ are privatized by clipping the per-sample gradients $\nabla_{s} \theta$ to an \textit{$L_2$-norm} threshold $C$ followed by addition of independent Gaussian noise with standard deviation $\sigma \times C$.
                \begin{equation}
                    \nabla_{s} \theta\leftarrow \nabla_s \theta) / \max \Big(1, \, \frac{||{\nabla_{s} \theta}||_2}{C}\Big)
                    \label{eq:clip}
                \end{equation}
                \begin{equation}
                    \nabla_b\theta \leftarrow \frac{1}{N} \big( \sum_{i=1}^{N}{\nabla_{s} \theta+ \mathcal{N}(0,\,\sigma C \textbf{I}_d)} \big)
                \end{equation}
            \end{theorem}
            Here $\sigma$ represents the noise multiplier.
        
         Recent work \cite{chen2020understanding} has shown that the clipping operation in Eq. \ref{eq:clip} creates geometric bias in the optimization trajectory of the loss landscape for DP-SGD. \cite{chen2020understanding} suggest to add Gaussian noise before clipping (referred to as \textit{pre-noising}) and prove that this helps to mitigate the geometric bias of the mini-batch gradients.
        
    \subsection{Stochastic Gradient Descent with Langevin Dynamics (SGLD)}
        Stochastic gradient Markov chain Monte Carlo (SG-MCMC) is part of a family of scalable Bayesian sampling algorithms that have recently (re)-emerged in the context of training of deep learning models on large datasets \cite{wenzel2020good}. In stochastic gradient Langevin dynamics (SGLD) \cite{welling2011bayesian}, Gaussian noise is added to the stochastic gradient descent update step. Through this addition of appropriately scaled noise proportional to the step size, the learning process converges to an MCMC chain and it is possible to draw samples from the \textit{posterior distribution} $p(\theta | D)$ over model parameters $\theta$. To guarantee convergence on this distribution, SGLD has two formal requirements: a decaying step size $\alpha$ and $0<\alpha<1$.
        
        To allow for differentially private learning without a direct and restrictive 1:1 coupling of the learning rate and noise multiplier, we use a \textit{temperature-scaled} reformulation of SGLD \cite{kim2020stochastic}.

    \begin{theorem}[\textbf{Temperature-scaled SGLD}]
        Given $x_N= \{x_1, x_2, ..., x_n\}$, a set of $n$ independent and identically distributed samples from the data distribution. Let $p(\theta_t)$ represent the prior distribution, $\alpha_t$ the step size at time $t$ and $U(\theta) =-\log p(\theta_t) - \log p(x_{N}|\theta_t))$ the energy function of the posterior. The SGLD update rule is then given by:
        \begin{align}
            \theta_{t+1} &= \theta_{t} - \alpha_{t} \nabla \widetilde{U}(\theta_{t}) + \eta_t\\
            \eta_t & \sim \mathcal{N}(0, \sqrt{2\alpha_t \tau}\mathbf{I}_d)
        \end{align}
        where $\tau$ represents the temperature parameter and $\nabla \widetilde{U}(\theta)$, a mini-batch estimate of $\nabla U(\theta)$. 
    \end{theorem}
    
    \subsection{Model Calibration}
        We use the definition of \cite{nixon2019measuringCalibration} for model calibration: Consider a dataset of data example pairs $\{(x_1, y_1), ...., (x_n, y_n)\}$ which we assume to be independent, identically distributed (i.i.d.) realizations of the random variables $X, Y \sim \mathbb{P}$.
        
        \begin{theorem}
            A model which predicts a class $y\in\{0, ...., K-1\}$ with probability $\hat{p}$ is well calibrated if and only if $\hat{p}$ is always the true probability.
            \begin{equation}
                \mathbb{P}\big(Y=y| \hat{p}=p\big)=p, \,\, \forall p\in[0,1]
                \label{eq:calibration}
            \end{equation}
            Where we define any difference between the right and left hand side of the above as calibration error.
        \end{theorem}
        
        Intuitively, this means, that a prediction from a well-calibrated model with 0.7 confidence should be correct 70\% of the time.
    
    \subsubsection{Expected Calibration Error}
    The expected calibration error (ECE) \cite{naeini2015obtaining}, partitions Eq. \ref{eq:calibration} into $M$ equally spaced bins and calculates the weighted average of the difference between confidence (maximum of softmax output) and accuracy, a lower ECE is better. ECE is defined as follows and approximates the calibration error for the most likely class:
    \begin{equation}
        ECE = \sum_{m=1}^{M}\frac{|B_m|}{n} | \mbox{acc.}(B_m)-\mbox{conf.}(B_m)|
        \label{eq:ECE}
    \end{equation}

\section{Differentially private stochastic gradient descent with Langevin dynamics (DP-SGLD)}
    To adapt the DP-SGD algorithm to the Bayesian setting and allow for sampling from the posterior distribution, we replace the noise multiplier term $\sigma$ in DP-SGD with a temperature-scaled multiple of the learning rate: $\sqrt{2\alpha_t \tau}$. The proposed changes are shown in Algorithm \ref{alg:dpsgld_algo}.
    
    \begin{algorithm}[h]
       \caption{DP-SGLD Algorithm}
       \label{alg:dpsgld_algo}
    \begin{algorithmic}
       \STATE {\bfseries Input:} Examples $\{x_1, ..., x_N\}$, Loss function $L(\theta)$.
        \STATE {\bfseries Params:} temperature $\mathbf{\tau}$, gradient clipping bound $C$, pre-noise scale $\mathbf{\rho}$, decaying learning rate $\mathbf{\alpha}_t$ and mini-batch size $m$
       \FOR{$t\in [T]$}
        \STATE Sample mini-batch $M$ from training data with probability $\frac{m}{N}$
        \FOR{each $x_i$ in $M$}
        \STATE \textbf{Compute per example gradient}
        \STATE $\mathbf{g_t(x_i)} \leftarrow \nabla_{\theta_t} L(\theta_t, x_i)$
        \STATE \textbf{Pre-noise gradients}
        \STATE $\mathbf{\hat{g_t}} \leftarrow \mathbf{g_t(x_i)} + \mathcal{N}(0, \mathbf{\rho}I_d$)
        \STATE \textbf{Clip gradient}
        \STATE $\mathbf{\overline{g}_t(x_i)} \leftarrow \mathbf{\hat{g_t}(x_i)} / \max \Big(1, \, \frac{||\mathbf{\hat{g_t}(x_i)}||_2}{\mathbf{C}}\Big)$
        \ENDFOR
        \STATE \textbf{Add noise}
        \STATE $\mathbf{\widetilde{g}_t} = \frac{1}{L} \sum_i{\mathbf{\overline{g}_t(x_i)} +\mathcal{N}(0,\,\sqrt{2\mathbf{\alpha}_t \mathbf{\tau}}C I_d) }$
        \STATE \textbf{Perform parameter update step}
        \STATE $\mathbf{\theta_t = \theta_{t-1} - \alpha_t \, \mathbf{\widetilde{g}_t}}$
       \ENDFOR
    \end{algorithmic}
    \end{algorithm}
    
    Note that, to satisfy the formal posterior convergence properties of SGLD, a decaying learning rate schedule is required.

\subsection{Privacy Analysis}
    \label{section:priv_anal}
    As Algorithm \ref{alg:dpsgld_algo} is equivalent to DP-SGD in the sense that the noise multiplier parameter is replaced by the learning rate and temperature to scale the standard deviation of the Gaussian noise, privacy accounting as originally proposed by \cite{abadi2016deep} remains unchanged, although we use use Gaussian differential privacy \cite{dong2019gaussian}, which offers a tighter bound on the privacy loss. Formally, the noise added to the stochastic gradient update step after clipping in DP-SGLD is distributed as follows:
    
    \begin{equation}
        \eta_{1} \sim \mathcal{N}(0, \sqrt{2\mathbf{\alpha}_t \mathbf{\tau}} C I_d)
        \label{eq:dpsgld_noise}
    \end{equation}
    and in DP-SGD:
    \begin{equation}
        \eta_{2} \sim \mathcal{N}(0, \sigma C I_d)
        \label{eq:dpsgd_noise}
    \end{equation}
    
    Thus it follows that $\eta_1$ and $\eta_2$ are equal if $\sigma=\sqrt{2\mathbf{\alpha}_t \mathbf{\tau}}$.

    \subsubsection{Privacy accounting with decaying noise}
    To perform privacy accounting with a decaying noise schedule we use the \textit{n-fold composition theorem} of Gaussian differential privacy \cite{dong2019gaussian}:
    
    \textit{For a sequence of $n$ $\mu_i$-GDP mechanisms composed over the dataset, the resulting mechanism is $\mu=\sqrt{\mu_1^2 + \mu_2^2 + ... +  \mu_n^2}$-GDP.}
    
    As a result, we can simply account for every optimization step with unique noise addition separately and aggregate the resulting $\mu$ values to calculate the final GDP privacy guarantees, which can then be converted to an $(\epsilon, \delta)$-DP.
        
\section{Experiments}
\begin{figure*}[ht!]
    \centering
    \includegraphics[width=0.9\textwidth]{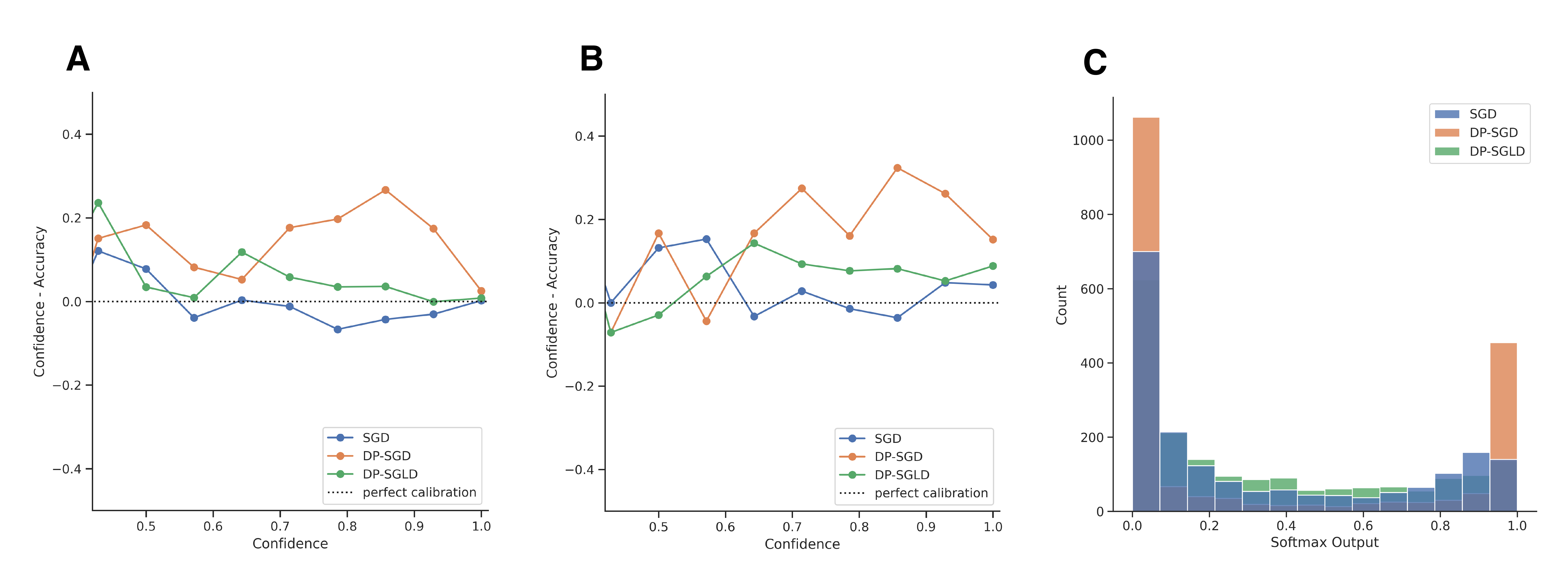}
    \vspace{-5mm}
    \caption{Calibration curves for \textit{MNIST} (A) and \textit{Pediatric Pneumonia} (B), alongside histogram plots of softmax output for \textit{Pediatric Pneumonia} (C), comparing \textit{SGD} (blue), \textit{DP-SGD} (orange) and \textit{DP-SGLD} (green). Confidence is the maximum value of the softmax output. Perfect calibration, defined as the equality of confidence and accuracy across the whole probability interval is shown by the dashed line. Points below the dashed line represent underconfident predictions, while points above represent overconfident predictions.}
    \label{fig:cal_curve}
    \vspace{-2mm}
\end{figure*}

We trained and tested our algorithm on two datasets (official train-test splits): \textit{MNIST} \cite{lecun1998gradient} and \textit{Pediatric Pneumonia dataset} \cite{kermany2018identifying}. We report the performance and calibration metrics for DPSGD, SGD and DP-SGLD (our algorithm). We note that the aim of our experiments was not to achieve new state-of-the-art results but to highlight calibration differences between DP-SGD and our proposed method. We report ECE as the only quantitative calibration metric as differences between ECE, SCE and ACE were negligible. For a visualisation of model calibration see the calibration curves in Fig. \ref{fig:cal_curve}.

\subsection{Results on MNIST}
Results for a five layer convolutional neural network (CNN) trained on MNIST with the compared optimization procedures. Once the privacy budget ($\epsilon=0.5$ at $\delta=1\times10^{-5}$) was exhausted, training was halted.

\begin{table}[h]
    \caption{Performance and calibration metrics for SGD, DP-SGD and DP-SGLD on MNIST.}
    \vspace{-5mm}
    \label{mnist-table}
    \begin{center}
    \begin{small}
    \begin{sc}
    \begin{adjustbox}{max width=0.5\textwidth}
    \begin{tabular}{lccccc}
        \toprule
        Procedure & \text{\large{$\epsilon$}} & Accuracy & AUC & \textbf{ECE} \\
        \midrule
        SGD & $\infty$ & 0.984 & 0.999 & \textbf{0.0020} \\
        DP-SGD & 0.5 &  0.967 & 0.996 & 0.0210  \\
        DP-SGLD (ours) & 0.5 & 0.963 & 0.995 & 0.0044 \\
        \bottomrule
    \end{tabular}
    \end{adjustbox}
    \end{sc}
    \end{small}
    \end{center}
    
\end{table}

\subsection{Results on Pediatric Pneumonia Dataset}
Results for a pre-trained, frozen backbone (EfficientNet B1 \cite{tan2019efficientnet}, ImageNet) with two trainable dense layers for the Pediatric Pneumonia dataset (PPD) \cite{kermany2018identifying} are shown in Table \ref{4P-table}. PPD is a multi-class classification dataset comprising chest radiographs (classes: bacterial pneumonia, viral pneumonia and normal). Once the privacy budget ($\epsilon=6.0$ at $\delta=1\times10^{-4}$) was exhausted, training was halted.

\begin{table}[h]
    \caption{Performance and calibration metrics for SGD, DP-SGD and DP-SGLD on the Pediatric Pneumonia dataset.}
    \vspace{-5mm}
    \label{4P-table}
    \begin{center}
    \begin{small}
    \begin{sc}
    \begin{adjustbox}{max width=0.5\textwidth}
    \begin{tabular}{lcccc}
    \toprule
    Procedure & \text{\large{$\epsilon$}} & Accuracy & AUC & \textbf{ECE} \\
    \midrule
    SGD & $\infty$ & 0.857 & 0.942 & \textbf{0.0526} \\
    DP-SGD & 6.0 & 0.783 & 0.910 & 0.1790 \\
    DP-SGLD (ours) & 6.0 & 0.786 & 0.920 & 0.0833 \\
    \bottomrule
    \end{tabular}
    \end{adjustbox}
    \end{sc}
    \end{small}
    \end{center}
\end{table}

\section{Discussion}
The proposed method provides substantially better calibrated predictions (MNIST $\sim{5}$-fold,  PPD $\sim{2}$-fold reduction in ECE) compared to standard DP-SGD. Furthermore, DP-SGLD provides uncertainty estimates with improved calibration across the ([0, 1]) model confidence interval as shown in Fig. \ref{fig:cal_curve} (A \& B). The softmax output of \textit{DP-SGD} is concentrated about the extreme values of the interval, indicating overconfidence for the \textit{Pediatric Pneumonia dataset}(Fig. \ref{fig:cal_curve} C). In contrast, the predicted probabilities of \textit{SGD} and \textit{DP-SGLD} were more homogeneously distributed.

Our work is not without limitations: DP-SGLD provides only point estimates of the network's parameters and thus sampling from the posterior when the privacy budget is exhausted is not possible. Furthermore, DP-SGLD converges onto a single mode in the posterior distribution, and is thus not capable of capturing multiple, diverse solutions (modes). Future work could explore other, more expressive probabilistic formulations of Bayesian neural networks such as DeepEnsembles \cite{lakshminarayanan2016simple} or SWAG \cite{maddox2019simple} for differentially private training. Our method employs pre-noising and we conjecture that this improves ergodicity in the posterior space, but leave it to future work to explore the effect of pre-noising and/or the decaying learning rate schedule on model calibration.

\section{Conclusion}
We present \textit{DP-SGLD}, an elegant reformulation of DP-SGD as Bayesian posterior inference and show that our approach yields much better calibrated models.


\bibliography{bibliography}
\bibliographystyle{icml2021}


\end{document}